# Domain-adaptive Message Passing Graph Neural Network

Xiao Shen[a], Shirui Pan[b], Kup-Sze Choi[c], Xi Zhou[d*]

[a] *School of Computer Science and Technology, Hainan University, Haikou, China*
[b] *School of ICT, Griffith University, Gold Coast, Australia*
[c] *Centre for Smart Health, The Hong Kong Polytechnic University, Hong Kong, China*
[d] *College of Tropical Crops, Hainan University, Haikou, China*

**Abstract**

Cross-network node classification (CNNC), which aims to classify nodes in a label-deficient target network by transferring the knowledge from a source network with abundant labels, draws increasing attention recently. To address CNNC, we propose a domain-adaptive message passing graph neural network (DM-GNN), which integrates graph neural network (GNN) with conditional adversarial domain adaptation. DM-GNN is capable of learning informative representations for node classification that are also transferrable across networks. Firstly, a GNN encoder is constructed by dual feature extractors to separate ego-embedding learning from neighbor-embedding learning so as to jointly capture commonality and discrimination between connected nodes. Secondly, a label propagation node classifier is proposed to refine each node's label prediction by combining its own prediction and its neighbors' prediction. In addition, a label-aware propagation scheme is devised for the labeled source network to promote intra-class propagation while avoiding inter-class propagation, thus yielding label-discriminative source embeddings. Thirdly, conditional adversarial domain adaptation is performed to take the neighborhood-refined class-label information into account during adversarial domain adaptation, so that the class-conditional distributions across networks can be better matched. Comparisons with eleven state-of-the-art methods demonstrate the effectiveness of the proposed DM-GNN.

**Key words:** Cross-network Node Classification, Domain Adaptation, Graph Neural Network, Message Passing

## 1. Introduction

Networks that model the relations between entities are ubiquitous in the real world, such as social networks, e-commerce networks, transportation networks, molecules networks and biological networks. Node classification is a common graph machine

* Corresponding author
Email address: shenxiaocam@163.com (Xiao Shen), s.pan@griffith.edu.au (Shirui Pan), thomasks.choi@polyu.edu.hk (Kup-Sze Choi), xzhou@hainanu.edu.cn (Xi Zhou)

learning task which aims to accurately classify unlabeled nodes in a network given a subset of labeled nodes (Bhagat et al., 2011). The existing node classification tasks mostly focus on a single-network scenario where the training nodes and testing nodes are all sampled from one single network (Kipf & Welling, 2017; Liang et al., 2018; Pan et al., 2016; Veličković et al., 2018; Yang et al., 2016; Zhang et al., 2018).

However, in practice, it is often resource and time-intensive to manually gather node labels for each newly formed target network, while abundant node labels might have already been accessible in some auxiliary networks. Cross-network node classification (CNNC) describes the problem of node classification across different networks with different distributions by transferring the knowledge from a relevant labeled source network to accurately classify unlabeled nodes in a target network (Dai et al., 2023; Shen, Dai, et al., 2020; Shen et al., 2021; M. Wu et al., 2020). CNNC is a valuable technique for a range of real-world applications. For example, in cross-network influence maximization, in order to maximize the influence in a new target network, CNNC can transfer the knowledge learned from a smaller source network, where all nodes have annotated labels reflecting their influence, to assist in the selection of the most influential nodes for the target network (Shen, Mao, et al., 2020). In cross-domain protein function prediction, given a source protein-protein interaction (PPI) network with abundant labels indicating protein functionalities, CNNC can help predict the protein functions in a new target PPI network (Hu et al., 2020). In cross-system recommendations, it is beneficial to transfer the knowledge learned from an online social network (e.g. Netflix) of users with plenty of social tags indicating their movie interests to predict the movie interests of users in another target online social network (e.g. Douban) (Zhu et al., 2018).

To succeed in CNNC, one need to address the challenges from two aspects, namely, the node classification problem and the cross-network problem. For the former, the challenge is how to integrate topological structures, node content information and observed node labels to learn informative representations for subsequent classification. For the latter, given the inherent different distributions of topologies and node content between the source and target networks, the key challenge is how to mitigate the domain discrepancy and yield network-transferable representations.

Graph neural networks (GNNs) (Z. Wu et al., 2020), which apply deep neural networks on graph-structured data, have become the state-of-the-art network representation learning method. GNNs typically adopt a message passing paradigm to aggregate the node's own features and the features of its neighbors to learn informative representations, which have demonstrated outstanding performance in semi-supervised node classification (Hamilton et al., 2017; Kipf & Welling, 2017; Liang et al., 2018; Pan et al., 2016; Veličković et al., 2018; Yang et al., 2016; Zhang et al., 2018). However, since the training nodes and testing nodes in CNNC are sampled from different networks with different distributions, the domain discrepancy across networks impede direct use of a GNN trained on a source network for a new target network (Pan & Yang, 2010). Thus, the existing GNNs fail to learn network-transferable representations given that they focus only on single-network representation learning while without considering domain discrepancy across different networks (Shen, Dai, et al., 2020; Shen et al., 2021).

Domain adaptation is an effective approach to mitigate the shift in data distributions across domains (Pan & Yang, 2010). Despite the significant achievements of domain adaptation in computer vision (CV) (Zhang, Li, et al., 2023; Zhang, Li, Tao, et al., 2021; Zhang, Li, Zhang, et al., 2021; Zhang et al., 2022; Zhang, Zhang, et al., 2023) and natural language processing (NLP) (Ramponi & Plank, 2020; Saunders, 2022), applying domain adaptation on graph-structured data is still non-trivial. This is because the conventional domain adaptation algorithms assume that every data sample in each domain is independent and identically distributed (i.i.d.). Clearly, graph-structured data obviously violate the i.i.d. assumption, i.e., nodes are not independent but relate to others via complex network connections. Therefore, the traditional domain adaptation algorithms based on the i.i.d. assumption would fail to model complex network structures and consequently perform poorly in CNNC (Shen, Dai, et al., 2020; Shen et al., 2021). However, it has been widely acknowledged that taking full advantage of network relationships between nodes should be indispensable to node classification (Hamilton et al., 2017; Kipf & Welling, 2017; Liang et al., 2018; Pan et al., 2016; Veličković et al., 2018; Wang et al., 2020; Yang et al., 2016; Zhang et al., 2018).

Given that GNNs are limited in handling domain discrepancy across networks and domain adaptation algorithms are limited in modeling complex network structures, utilizing either GNNs or domain adaptation alone cannot effectively tackle the challenges in the CNNC problem. To go beyond such limits, the integration of GNN with domain adaptation has become a promising paradigm to address CNNC (Dai et al., 2023; Shen, Dai, et al., 2020; Shen et al., 2021; M. Wu et al., 2020; Zhang et al., 2019). However, the existing CNNC algorithms have the two main weaknesses that need to be addressed but remain neglected. Firstly, the algorithms (Dai et al., 2023; M. Wu et al., 2020; Zhang et al., 2019) typically adopt graph convolutional network (GCN) (Kipf & Welling, 2017) or its variants (Li et al., 2019; Zhuang & Ma, 2018) for node embedding learning, while the entanglement of neighborhood aggregation and representation transformation in the GCN-like models can easily leads to over-smoothing (Dong et al., 2021; Klicpera et al., 2019; Li et al., 2018; Liu et al., 2020). In addition, in the GCN-like models, the ego-embedding and neighbor-embedding of each node are mixed at each graph convolution layer, the discrimination between connected nodes is not preserved, which yields poor classification performance when the connected nodes are dissimilar (Bo et al., 2021; Luan et al., 2020; Zhu et al., 2020). Secondly, the existing CNNC algorithms (Dai et al., 2023; Shen, Dai, et al., 2020; M. Wu et al., 2020; Zhang et al., 2019) mainly focus on matching the marginal distributions across networks but neglect the class-conditional shift across networks which can significantly hamper the finding of a joint ideal classifier for the source and target data (Long et al., 2018; Luo et al., 2020).

To remedy the aforementioned weaknesses, we propose a novel d̲omain-adaptive m̲essage passing g̲raph n̲eural network (DM-GNN) to address CNNC. Firstly, to tackle the limitations of the GCN-like models which are widely adopted in existing CNNC algorithms (Dai et al., 2023; M. Wu et al., 2020; Zhang et al., 2019), a GNN encoder is constructed by dual feature extractors with different learnable parameters to learn the ego-embedding and neighbor-embedding of each node respectively. As a result, both the commonality and discrimination between connected nodes can be effectively captured. In addition, unlike the GCN-like models, the neighborhood aggregation and representation transformation have been decoupled by a second feature extractor to alleviate the over-smoothing problem. Secondly, a feature propagation loss is proposed to smooth the embeddings w.r.t. graph topology by encouraging each node's final embedding to become similar to a weighted average of its neighbors' final embeddings. Thirdly, most existing CNNC algorithms (Shen, Dai, et al., 2020; Shen et al., 2021; M. Wu et al., 2020; Zhang et al., 2019) only focus on neighborhood aggregation at feature-level, and simply adopt a logistic regression or a multi-layer perceptron (MLP) as the node classifier. In contrast, DM-GNN unifies feature propagation with label propagation by proposing a label propagation mechanism in the node classifier to smooth the label prediction among the neighborhood. As a result, the label prediction of each node can be refined by combining its own prediction and the prediction from its neighbors within $K$ steps. Moreover, for the fully labeled source network, we devise a label-aware propagation scheme to only allow the message (i.e. both features and label prediction) passing through the same labeled neighbors, which promotes intra-class propagation while avoiding inter-class propagation. Thus, label-discriminative source embeddings can be learned by DM-GNN. Lastly, instead of only matching the marginal distributions across networks in most existing CNNC algorithms (Dai et al., 2023; Shen, Dai, et al., 2020; M. Wu et al., 2020; Zhang et al., 2019), a conditional adversarial domain adaptation approach (Long et al., 2018) is employed by DM-GNN to align the class-conditional distributions between the source and target networks. Specifically, a conditional domain discriminator is adopted to consider both embeddings and label prediction during adversarial domain adaptation. Instead of directly utilizing the own label prediction of each sample independently as in (Long et al., 2018), empowered by the label propagation node classifier, the proposed DM-GNN can utilize the neighborhood-refined label prediction during conditional adversarial domain adaptation, which guarantees more accurate predicted labels available to guide the alignment of the corresponding class-conditional distributions between two networks. With the label-discriminative source embeddings and class-conditional adversarial domain adaptation, the target nodes would be aligned to the source nodes with the same class to yielding label-discriminative target embeddings.

The main contributions are summarized as follows:

1) To tackle the problems with the GCN-like models in most existing CNNC algorithms, a simple and effective GNN encoder with dual feature extractors is employed to separate ego-embedding from neighbor-embedding, and thereby jointly capture the commonality and discrimination between connected nodes. In addition, through decoupling neighborhood aggregation from representation transformation with a second feature extractor, the over-smoothing issue with the GCN-like models can be alleviated.

2) The proposed DM-GNN performs message passing among the neighborhood at both feature level and label level by designing a feature propagation loss and a label propagation node classifier respectively. A label-aware propagation mechanism is devised for the fully labeled source network to promote intra-class propagation while avoiding inter-class propagation to guarantee more label-discriminative source embeddings. In addition, a conditional adversarial domain adaptation approach is employed by DM-GNN to consider the neighborhood-refined label prediction during adversarial domain adaptation to align the target embeddings to the source embeddings associated with the same class.

3) Extensive experiments conducted on the benchmark datasets demonstrate the outperformance of the proposed DM-GNN for CNNC over eleven state-of-the-art methods, and the ablation study verifies the effectiveness of the model designs.

## 2. Related Works

*2.1 Network Representation Learning*

Network embedding is an effective network representation learning method to learn low-dimensional representations for graph-structured data. Pioneering network embedding methods, however, only consider plain network structures (Dai et al., 2018; Dai et al., 2019; Grover & Leskovec, 2016; Liu et al., 2021; Perozzi et al., 2014; Shen & Chung, 2017, 2020; Wang et al., 2020). Other than plain network structures, attributed network embedding algorithms further take node content information and available labels into account to learn more meaningful representations. GNNs are among the top-performing attributed network embedding algorithms, which refine the representation of each node by combining the information from its neighbors. Graph Convolutional Network (GCN) (Kipf & Welling, 2017) is the most representative GNN, which employs a layer-wise propagation rule to iteratively aggregate node features through the edges of the graph. Many follow-up GNNs inspired by GCN have been proposed. For example, Velickovic *et al.* (Veličković et al., 2018) developed a GAT to leverage an attention mechanism to automatically learn appropriate edge weights during neighborhood aggregation. Hamilton *et al.* (Hamilton et al., 2017) proposed a GraphSAGE to support several neighborhood aggregation methods beyond averaging. Chen *et al.* (Chen et al., 2020) proposed a label-aware GCN to filter negative neighbors with different labels and add new positive neighbors with same labels during neighborhood aggregation. Yang *et al.* (Yang et al., 2020) developed a PGCN model which integrates network topology and node content information from a probabilistic perspective.

In the GCN-like models, two important operations, i.e., feature transformation and neighborhood aggregation, are typically entangled at each convolutional layer. Recent studies showed that such entanglement is unnecessary and would easily lead to over-smoothing (Dong et al., 2021; Klicpera et al., 2019; Li et al., 2018; Liu et al., 2020). In addition, in typical GCN design, ego-embedding and neighbor-embedding are mixed through the average or weighted average aggregation method at each convolutional layer (Zhu et al., 2020). The mixing design performs well when the connected nodes have the same label but results in poor performance when the connected nodes have dissimilar features and different labels (Bo et al., 2021; Luan et al., 2020; Zhu et al., 2020).

Although GNNs have achieved impressive performance in semi-supervised node classification, they are generally developed under a single-network setting, without considering domain discrepancy across different networks. Therefore, they fail to tackle prediction tasks across different networks effectively (Heimann et al., 2018; Shen et al., 2021).

*2.2 Domain Adaptation*

Domain adaptation is an effective approach to mitigate data distribution shift across domains. Self-training domain adaptation approaches (Chen et al., 2011; Shen et al., 2017; Shen, Mao, et al., 2020) are proposed to iteratively add the target samples with high prediction confidence into the training set. Feature-based deep domain adaptation algorithms have drawn attentions recently, which can be categorized into two families, i.e., using statistical approaches and adversarial learning approaches respectively. The family using statistical approaches reduces domain shift by minimizing the statistical metrics which measure the distribution discrepancy between different domains, such as Maximum Mean Discrepancy (MMD) (Gretton et al., 2007) and conditional MMD (Long et al., 2013). Inspired by Generative Adversarial Networks (GAN) (Goodfellow et al., 2014), a family of adversarial domain adaptation algorithms is developed to minimize domain shift by training a generator and a domain discriminator in an adversarial manner (Ganin et al., 2016; Shen et al., 2018; Tzeng et al., 2017).

Most existing adversarial domain adaptation algorithms just focus on matching the marginal distributions across domains. However, matching the marginal distribution does not guarantee that the corresponding class-conditional distributions can be well matched. It has been revealed that conditional GANs (Mirza & Osindero, 2014; Odena et al., 2017) which condition the generator and discriminator on the associated label information can better align different distributions. Motivated by this finding, several conditional adversarial domain adaptation algorithms taking class-conditional information into account during adversarial domain adaptation have been proposed. For example, Long *et al.* (Long et al., 2018) proposed a CDAN model which conditions adversarial domain adaptation on the discriminative label information predicted by the task classifier. Pei *et al.* (Pei et al., 2018) proposed a MADA approach to employ multiple class-wise domain discriminators to capture the multimodal structures of data distributions across domains. Tang and Jia (Tang & Jia, 2020) proposed a DADA model to align the joint distributions of feature and category across domains, by adopting an integrated classifier which jointly predicts the category and domain labels.

The domain adaptation algorithms developed in CV and NLP are typically based on the i.i.d. assumption. Nevertheless, due to the non-i.i.d. nature of graph-structured data, the traditional domain adaptation algorithms are limited in their capabilities to tackle the prediction tasks across networks (Dai et al., 2023; Shen, Dai, et al., 2020; Shen et al., 2021; M. Wu et al., 2020).

*2.3 Cross-network Node Classification*

CNNC draws increasing attention very recently. It aims to transfer the knowledge from a source network with abundant labeled data to accurately classify the nodes in a target network where labels are lacking. Shen *et al.* (Shen et al., 2021) proposed a CDNE model to integrate two stacked auto-encoders (SAEs) with the MMD and conditional MMD metrics. Zhang *et al.* (Zhang et al., 2019) proposed a DANE model which applies GCN and adversarial domain adaptation to learn transferable embeddings in an unsupervised manner. Wu *et al.* (M. Wu et al., 2020) proposed a UDAGCN model which employs adversarial domain adaptation and a dual GCN framework (Zhuang & Ma, 2018) to capture the local consistency and global consistency of the graphs. Shen *et al.* (Shen, Dai, et al., 2020) proposed an ACDNE model to integrate adversarial domain adaptation with dual feature extractors which learn the representation of each node separately from that of its neighbors. Dai *et al.* (Dai et al., 2023) proposed an AdaGCN model to integrate GCN with adversarial domain adaptation. Besides, an AdaIGCN model is further proposed, which employs an improved GCN layer (Li et al., 2019) to flexibly adjust the neighborhood size in feature smoothing.

Most existing CNNC algorithms (Shen, Dai, et al., 2020; Shen et al., 2021; M. Wu et al., 2020; Zhang et al., 2019) focus on feature propagation. The DM-GNN proposed in this paper unifies feature propagation and label propagation in cross-network embedding, which incorporates a label propagation mechanism in the node classifier to refine each node's label prediction by combining the label predictions from its neighbors. Moreover, for the labeled source network, a label-aware propagation mechanism is devised to promote intra-class propagation while avoiding inter-class propagation to produce more label-discriminative source embeddings. In addition, the adversarial domain adaptation approaches employed in most existing CNNC algorithms (Dai et al., 2023; Shen, Dai, et al., 2020; M. Wu et al., 2020; Zhang et al., 2019) only focus on matching the marginal distributions across networks, which cannot guarantee the corresponding class-conditional distributions across networks to be well aligned. In contrast, the proposed DM-GNN employs the conditional adversarial domain adaptation approach to condition GNN and domain discriminator on the neighborhood-refined class-label information. As a result, nodes from different networks but having the same class-label would have similar representations.

## 3. Proposed Model

In this section, we describe in detail the three components of the proposed DM-GNN, namely, dual feature extractors, a label propagation node classifier and a conditional domain discriminator. The model architecture of DM-GNN is shown in Fig. 1.

*3.1 Notations*

Let $\mathcal{G} = (V, E, \mathbf{A}, \mathbf{X}, \mathbf{Y})$ denote a network with a set of nodes $V$, a set of edges $E$, a topological proximity matrix $\mathbf{A} \in \mathbb{R}^{\mathcal{N} \times \mathcal{N}}$, a node attribute matrix $\mathbf{X} \in \mathbb{R}^{\mathcal{N} \times \mathcal{W}}$, and a node label matrix $\mathbf{Y} \in \mathbb{R}^{\mathcal{N} \times \mathcal{C}}$, where $\mathcal{N}, \mathcal{W}$ and $\mathcal{C}$ are the number of nodes, node attributes and node labels in $\mathcal{G}$ respectively. Each node $v_i \in V$ is associated with a topological proximity vector $\mathbf{a}_i$, an attribute vector $\mathbf{x}_i$, and a label vector $\mathbf{y}_i$. Specifically, $\mathbf{y}_i$ is a one-hot vector if $v_i$ is a labeled node; otherwise, $\mathbf{y}_i$ is a zero vector. The superscripts $s$ and $t$ are employed to denote the source network and the target network. In this work, we focus on unsupervised CNNC problem, i.e., all nodes in the source network $\mathcal{G}^s = (V^s, E^s, \mathbf{A}^s, \mathbf{X}^s, \mathbf{Y}^s)$ have observed labels and all nodes in the target network $\mathcal{G}^t = (V^t, E^t, \mathbf{A}^t, \mathbf{X}^t)$ are unlabeled. Note that $\mathcal{G}^s$ is structurally very different from $\mathcal{G}^t$, and the attribute distributions of $\mathcal{G}^s$ and $\mathcal{G}^t$ are also very different. The goal of CNNC is to learn appropriate cross-network embeddings based upon which a node classifier trained on the source network can be applied to accurately classify nodes for the target network. For clarity, Table 1 summarizes the frequently used notations in this article.

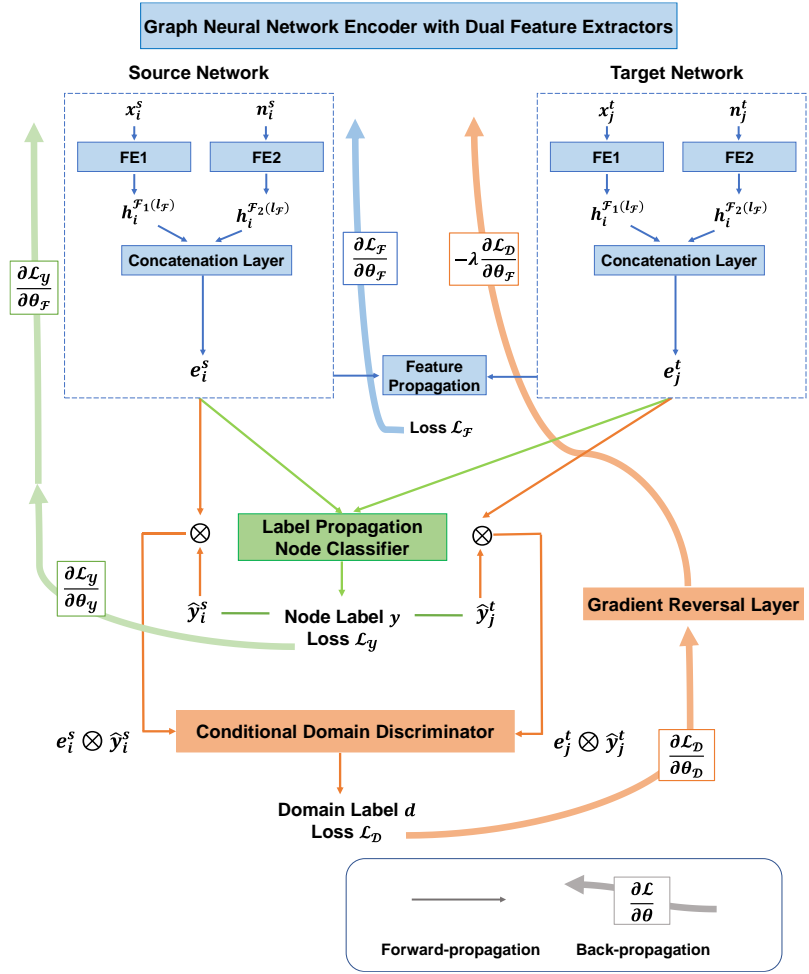

Fig. 1. Model architecture of DM-GNN. A GNN encoder with dual feature extractors is employed to separate ego-embedding learning from neighbor-embedding learning. A label propagation node classifier is employed to refine label prediction. A conditional domain discriminator taking embeddings and predicted labels as the input is employed to compete against the GNN encoder by inserting a gradient reversal layer during back-propagation.

*3.2 Graph Neural Network Encoder with Dual Feature Extractors*

GCN-like models have been widely adopted by the state-of-the-art CNNC algorithms (Dai et al., 2023; M. Wu et al., 2020; Zhang et al., 2019) for network representation learning. In these models, ego-embedding and neighbor-embedding are typically mixed at each convolutional layer. Then, the same learnable parameters are employed on such mixed embeddings for representation transformation. It has been revealed in (Bo et al., 2021; Luan et al., 2020; Zhu et al., 2020) that the GCN-like models fail to preserve the discrimination between connected nodes since the layer-wise mixing strategy always forces connected nodes to have similar representations, even though they possess dissimilar attributes. In addition, recent literatures (Dong et al., 2021; Klicpera et al., 2019; Liu et al., 2020) have also revealed that the entanglement of neighborhood aggregation and representation transformation can lead to over-smoothing easily, eventually making the nodes from different classes having indistinguishable representations.

To go beyond the limits of the typical GCN design, we construct a novel GNN encoder with dual feature extractors, which is distinct from the GCN-like models. Instead of mixing each node's embedding with its neighbors' embeddings, dual feature extractors with different learnable parameters are adopted to separate ego-embedding learning from neighbor-embedding learning. The two feature extractors are both constructed by an MLP with the same layer setting. However, the input of the two feature extractors is different. The first feature extractor (FE1) is employed to learn ego-embedding, hinging on each node's own attributes, as:

$$\boldsymbol{h}_i^{\mathcal{F}_1(l)} = \text{ReLU}\left(\boldsymbol{h}_i^{\mathcal{F}_1(l-1)} \boldsymbol{W}^{\mathcal{F}_1(l)} + \boldsymbol{b}^{\mathcal{F}_1(l)}\right), 1 \leq l \leq l_\mathcal{F} \tag{1}$$

where $\boldsymbol{h}_i^{\mathcal{F}_1(0)} = \boldsymbol{x}_i$, and $\boldsymbol{h}_i^{\mathcal{F}_1(l)}$ represents the *l*-th layer ego-embedding of $v_i$. $\boldsymbol{W}^{\mathcal{F}_1(l)}$ and $\boldsymbol{b}^{\mathcal{F}_1(l)}$ are the trainable parameters associated with the *l*-th layer of FE1, and $l_\mathcal{F}$ is the number of hidden layers in FE1.

In addition, we perform feature propagation to aggregate the neighbor attributes of each node as follows:

$$\boldsymbol{n}_i = \sum_{j=1, j \neq i}^{\mathcal{N}} \frac{a_{ij}}{\sum_{g=1, g \neq i}^{\mathcal{N}} a_{ig}} \boldsymbol{x}_j \tag{2}$$

where $\boldsymbol{n}_i \in \mathbb{R}^{1 \times \mathcal{W}}$ is the aggregated neighbor attribute vector of $v_i$. $a_{ij}$ is the topological proximity between $v_i$ and $v_j$, which is measured by the positive pointwise mutual information (PPMI) metric (Levy & Goldberg, 2014). The PPMI metric was originally developed to measure the similarity between words in NLP. It was extended to measure high-order proximities between nodes in random-walk-based network embedding algorithms (Cao et al., 2016; Perozzi et al., 2014; Shen & Chung, 2017). It has been widely acknowledged that the PPMI metric which captures the high-order proximities is effective for node classification (Cao et al., 2016; Perozzi et al., 2014; Shen & Chung, 2017), thus, it has been widely adopted by the state-of-the-art CNNC algorithms (Dai et al., 2023; Shen, Dai, et al., 2020; Shen et al., 2021; M. Wu et al., 2020) to measure topological proximities. Firstly, we follow (Shen & Chung, 2017) to compute an aggregated transition probability matrix $\mathcal{T}$ within $K$ steps, by assigning lower weights to more distant neighbors as $\mathcal{T} = \sum_{k=1}^{K} \mathcal{T}^{(k)}/k$, where $\mathcal{T}^{(k)}$ is the $k$-step transition probability matrix. Then, the topological proximity between $v_i$ and $v_j$ is measured by PPMI (Levy & Goldberg, 2014) as:

$$a_{ij} = \begin{cases} \max\left(\log\left(\frac{\mathcal{T}_{ij}/\sum_{g=1}^{\mathcal{N}} \mathcal{T}_{ig}}{\sum_{g=1}^{\mathcal{N}}(\mathcal{T}_{gj}/\sum_{l=1}^{\mathcal{N}} \mathcal{T}_{gl})/\mathcal{N}}\right), 0\right), if\ \mathcal{T}_{ij} > 0 \\ 0, if\ \mathcal{T}_{ij} = 0 \end{cases} \quad (3)$$

where $\mathcal{T}_{ij}$ is the aggregated transition probability from $v_i$ to $v_j$ within $K$ steps. Note that $a_{ij} > 0$ if $v_j$ can reach $v_i$ within $K$ steps in $\mathcal{G}$; otherwise, $a_{ij} = 0$.

Table 1
Frequently used notations.

| Notations | Descriptions |
|---|---|
| $\mathcal{G}$ | A network |
| $\mathcal{N}$ | Number of nodes in $\mathcal{G}$ |
| $\mathcal{W}$ | Number of attributes in $\mathcal{G}$ |
| $\mathcal{C}$ | Number of labels in $\mathcal{G}$ |
| $v_i$ | *i*-th node in $\mathcal{G}$ |
| $\boldsymbol{a}_i$ | Topological proximity vector of $v_i$ |
| $\boldsymbol{x}_i$ | Attribute vector of $v_i$ |
| $\boldsymbol{n}_i$ | Neighbor attribute vector of $v_i$ |
| $\boldsymbol{y}_i$ | Label vector of $v_i$ |
| $\boldsymbol{h}_i^{\mathcal{F}_1(l_\mathcal{F})}$ | Ego-embedding of $v_i$ learned by FE1 |
| $\boldsymbol{h}_i^{\mathcal{F}_2(l_\mathcal{F})}$ | Neighbor-embedding of $v_i$ learned by FE2 |
| $\boldsymbol{e}_i$ | Final embedding of $v_i$ |
| $\mathbb{d}$ | Embedding dimensionality |

The second feature extractor (FE2) is employed to learn neighbor-embedding, hinging on the aggregated neighbor attributes, as:

$$\boldsymbol{h}_i^{\mathcal{F}_2(l)} = \text{ReLU}\left(\boldsymbol{h}_i^{\mathcal{F}_2(l-1)} \boldsymbol{W}^{\mathcal{F}_2(l)} + \boldsymbol{b}^{\mathcal{F}_2(l)}\right), 1 \leq l \leq l_{\mathcal{F}} \quad (4)$$

where $\boldsymbol{h}_i^{\mathcal{F}_2(0)} = \boldsymbol{n}_i$, and $\boldsymbol{h}_i^{\mathcal{F}_2(l)}$ represents the $l$-th layer neighbor-embedding of $v_i$. $\boldsymbol{W}^{\mathcal{F}_2(l)}$ and $\boldsymbol{b}^{\mathcal{F}_2(l)}$ are the trainable parameters at the $l$-th layer of FE2. Note that in FE2, the neighborhood aggregation in Eq. (2) has been decoupled from the representation transformation in Eq. (4), which enable the number of propagation steps ($K$) to be independent of the number of hidden layers ($l_{\mathcal{F}}$) for representation transformation. It has been shown in (Dong et al., 2021; Klicpera et al., 2019; Liu et al., 2020) that such decoupling design can remedy over-smoothing in the GCN-like models.

Next, we formulate the deepest ego-embedding and deepest neighbor-embedding, and feed them to a single-layer perceptron for non-linear transformation, as follows:

$$\boldsymbol{e}_i = \text{ReLU}\left([\boldsymbol{h}_i^{\mathcal{F}_1(l_{\mathcal{F}})} \| \boldsymbol{h}_i^{\mathcal{F}_2(l_{\mathcal{F}})}] \boldsymbol{W}_c + \boldsymbol{b}_c\right) \quad (5)$$

where $\boldsymbol{e}_i \in \mathbb{R}^{1 \times \mathbb{d}}$ is the final embedding of $v_i$, $\mathbb{d}$ is the embedding dimensionality, $\boldsymbol{W}_c$ and $\boldsymbol{b}_c$ are the trainable parameters. Note that in contrast to the GCN-like models, we employ different trainable parameters to learn ego-embedding and neighbor-embedding separately. With the help of this separation design, the final embedding learned by DM-GNN is capable of capturing both commonality and discrimination between connected nodes:

(i) When two connected nodes $v_i$ and $v_j$ have similar attributes, FE1 generates similar ego-embeddings and FE2 generates similar neighbor-embeddings. By combining similar ego-embeddings and similar neighbor-embeddings, $v_i$ and $v_j$ would have similar final embeddings, which effectively capture the commonality between connected nodes.

(ii) When two connected nodes $v_i$ and $v_j$ have dissimilar attributes, FE2 still generates similar neighbor-embeddings due to network connection, whereas FE1 generates dissimilar ego-embeddings due to dissimilar attributes, thus the discrimination between connected nodes can be captured. Then, by concatenating the similar neighbor-embeddings and dissimilar ego-embeddings, the final embeddings of $v_i$ and $v_j$ would not become prohibitively similar.

We essentially conduct feature propagation in Eq. (2), i.e., propagating the input attributes of each node to its neighbors. Apart from propagating features during the input process, we further introduce a feature propagation loss to ensure that the output embeddings are also satisfied with the feature propagation objective, as:

$$\frac{1}{\mathcal{N}}\sum_{i=1}^{\mathcal{N}} \left\| e_i - \sum_{j=1,j\neq i}^{\mathcal{N}} \frac{a_{ij}}{\sum_{g=1,g\neq i}^{\mathcal{N}} a_{ig}} e_j \right\|^2 \tag{6}$$

Minimizing Eq. (6) makes each node's final embedding similar to a weighted average of its neighbors' final embeddings. Nevertheless, the aggregation in Eq. (6) does not consider class-label information. When a node has connections between different labeled nodes, aggregating the information from such different labeled neighbors would introduce noises and degrade the node classification performance. To avoid such noises, for the source network where all nodes have observed labels, we devise a label-aware feature propagation loss to allow features to propagate through the same labeled neighbors only, as:

$$\frac{1}{\mathcal{N}^s}\sum_{i=1}^{\mathcal{N}^s} \left\| e_i^s - \sum_{j=1,j\neq i}^{\mathcal{N}^s} \frac{a_{ij}^s o_{ij}^s}{\sum_{g=1,g\neq i}^{\mathcal{N}^s} a_{ig}^s o_{ig}^s} e_j^s \right\|^2 \tag{7}$$

where $o_{ij}^s$ is a label indicator showing whether $v_i^s$ and $v_j^s$ share common labels or not, i.e., $o_{ij}^s=1$ if $v_i^s$ and $v_j^s$ share at least one common label; otherwise, $o_{ij}^s=0$. Note that $a_{ij}^s o_{ij}^s>0$ if and only if $v_j^s$ is a same-labeled neighbor of $v_i^s$ within $K$ steps; otherwise, $a_{ij}^s o_{ij}^s=0$. Minimizing Eq. (7) constraints each source node to have similar embedding with its same-labeled neighbors, rather than with all of its neighbors. On the other hand, for the target network which has no label information, we use the feature propagation loss in Eq. (6). The total feature propagation loss of DM-GNN is defined as:

$$\mathcal{L}_\mathcal{F} = \frac{1}{\mathcal{N}^s}\sum_{i=1}^{\mathcal{N}^s} \left\| e_i^s - \sum_{j=1,j\neq i}^{\mathcal{N}^s} \frac{a_{ij}^s o_{ij}^s}{\sum_{g=1,g\neq i}^{\mathcal{N}^s} a_{ig}^s o_{ig}^s} e_j^s \right\|^2 + \frac{1}{\mathcal{N}^t}\sum_{i=1}^{\mathcal{N}^t} \left\| e_i^t - \sum_{j=1,j\neq i}^{\mathcal{N}^t} \frac{a_{ij}^t}{\sum_{g=1,g\neq i}^{\mathcal{N}^t} a_{ig}^t} e_j^t \right\|^2 \tag{8}$$

where $\mathcal{N}^s$ and $\mathcal{N}^t$ are the number of nodes in $\mathcal{G}^s$ and $\mathcal{G}^t$.

*3.3 Label Propagation Node Classifier*

Most existing CNNC algorithms (Shen, Dai, et al., 2020; Shen et al., 2021; M. Wu et al., 2020; Zhang et al., 2019) only focus on feature propagation, i.e., smoothing features of the neighboring nodes. The prior label propagation algorithms (Bengio et al., 2006; Zhou et al., 2004) aim to propagate the label probability distributions through the edges of the graph. Both GNNs and label propagation algorithms can be viewed as message passing algorithms on the graph, with the goal of feature smoothing and label smoothing over the neighborhood respectively. In DM-GNN, we propose to unify feature propagation and label propagation in cross-network embedding. Specifically, we incorporate label propagation in the node classifier by combining its own prediction and the prediction from its neighbors as follows:

$$\hat{y}_i = \phi\left(e_i W_y + b_y + \sum_{j=1, j \neq i}^{\mathcal{N}} \frac{a_{ij}}{\sum_{g=1, g \neq i}^{\mathcal{N}} a_{ig}} \left(e_j W_y + b_y\right)\right) \tag{9}$$

where $\hat{y}_i \in R^{1 \times C}$ is the predicted label probability vector of $v_i$. $W_y$ and $b_y$ are trainable parameters, and $e_i W_y + b_y$ is a vector of label logits (raw unnormalized label prediction) of $v_i$. $\phi(\cdot)$ can be a Softmax or Sigmoid function. In Eq. (9), the label logits of the neighbors which can reach $v_i$ within $K$ steps in $\mathcal{G}$ are aggregated. In addition, during label aggregation, higher weights are assigned to more closely connected neighbors (i.e. associated with higher topological proximities). Besides, similar to the label-aware feature propagation in Eq. (7), we modify Eq. (9) by incorporating the label indicator for the fully labeled source network as follows:

$$\hat{y}_i^s = \phi\left(\begin{array}{c} e_i^s W_y + b_y \\ + \sum_{j=1, j \neq i}^{\mathcal{N}^s} \frac{a_{ij}^s o_{ij}^s}{\sum_{g=1, g \neq i}^{\mathcal{N}^s} a_{ig}^s o_{ig}^s} \left(e_j^s W_y + b_y\right) \end{array}\right) \tag{10}$$

which only allows label propagation from the same labeled neighbors of $v_i^s$ and avoids label propagation from the neighbors having different labels with $v_i^s$. The cross-entropy node classification loss is defined as:

$$\mathcal{L}_y = \text{Cross Entropy}(\hat{y}_i^s, y_i^s) \tag{11}$$

where $y_i^s \in R^{1 \times C}$ is the ground-truth label vector of $v_i^s$. With the label-aware propagation mechanism in Eq. (10) and by minimizing Eq. (11), more label-discriminative source embeddings can be learned.

*3.4 Conditional Adversarial Domain Adaptation*

Inspired by GAN, a family of adversarial domain adaptation algorithms has been proposed, which demonstrate impressive performance in learning domain-invariant representations. The adversarial domain adaptation approaches (Ganin et al., 2016; Mao et al., 2017; Shen et al., 2018) adopted by the existing CNNC algorithms (Dai et al., 2023; Shen, Dai, et al., 2020; M. Wu et al., 2020; Zhang et al., 2019) focus on matching the marginal distributions of feature representations across networks, but adapting only feature representations cannot guarantee that the corresponding class-conditional distributions can be well matched (Long et al., 2018). To address this issue, in DM-GNN, we propose to condition adversarial domain adaptation on the class-label information. A simple solution is to directly concatenate the feature representation and label prediction as the input of the conditional domain discriminator (Mirza & Osindero, 2014; Odena et al., 2017). However, the concatenation strategy makes feature representation and label prediction independent of each other during adversarial domain adaptation (Long et al., 2018). To capture the multiplicative interactions between features and labels, we employ the tensor product between the embedding and label prediction as the input of conditional domain discriminator, as in the recent conditional adversarial domain adaptation algorithms (Long et al., 2018; Pei et al., 2018; Wang et al., 2019). Specifically, an MLP is employed to construct the conditional domain discriminator, i.e.,

$$\boldsymbol{h}_i^{\mathcal{D}(l)} = \text{ReLU}\big(\boldsymbol{h}_i^{\mathcal{D}(l-1)}\boldsymbol{W}^{\mathcal{D}(l)} + \boldsymbol{b}^{\mathcal{D}(l)}\big), 1 \leq l \leq l_{\mathcal{D}} \tag{12}$$

where $\boldsymbol{W}^{\mathcal{D}(l)}$ and $\boldsymbol{b}^{\mathcal{D}(l)}$ are trainable parameters, $l_{\mathcal{D}}$ is the number of hidden layers in conditional domain discriminator. $\boldsymbol{h}_i^{\mathcal{D}(0)} = \boldsymbol{e}_i \otimes \hat{\boldsymbol{y}}_i$ represents the input of conditional domain discriminator, $\otimes$ denotes the tensor product operation. $\hat{\boldsymbol{y}}_i$ is the neighborhood-refined label probability vector of $v_i$, which is predicted via Eq. (10) if $v_i$ is from the source network, or via Eq. (9) if $v_i$ is from the target network. Note that unlike the studies in (Long et al., 2018; Pei et al., 2018; Wang et al., 2019) that utilize the own label prediction of each sample independently, the neighborhood-refined label prediction proposed by DM-GNN is expected to yield more accurate label prediction during conditional adversarial domain adaptation. By utilizing $\boldsymbol{e}_i \otimes \hat{\boldsymbol{y}}_i$ as the input to conditional domain discriminator, the multimodal structures of the data distributions across networks can be well captured. In addition, the joint distribution of the embedding and class can be captured during adversarial domain adaptation. Next, a Softmax layer is added to predict which network a node comes from:

$$\hat{d}_i = \text{Softmax}\big(\boldsymbol{h}_i^{\mathcal{D}(l_{\mathcal{D}})}\boldsymbol{W}^{\mathcal{D}} + \boldsymbol{b}^{\mathcal{D}}\big) \tag{13}$$

where $\hat{d}_i$ is the probability of $v_i$ coming from the target network, predicted by the conditional domain discriminator. $\boldsymbol{W}^{\mathcal{D}}$ and $\boldsymbol{b}^{\mathcal{D}}$ are trainable parameters. By employing nodes from the two networks for training, the cross-entropy domain classification loss is defined as:

$$\mathcal{L}_{\mathcal{D}} = \underset{v_i \in \{V^s \cup V^t\}}{\text{Cross Entropy}}\big(\hat{d}_i, d_i\big) \tag{14}$$

where $d_i$ is the ground-truth domain label of $v_i$, $d_i = 1$ if $v_i \in V^t$ and $d_i = 0$ if $v_i \in V^s$.

*3.5 Optimization of DM-GNN*

The DM-GNN is trained in an end-to-end manner, by optimizing the following minmax objective function:

$$\min_{\theta_{\mathcal{F}},\theta_y} \left\{\mathcal{L}_y + f\mathcal{L}_{\mathcal{F}} + \lambda \max_{\theta_{\mathcal{D}}}\{-\mathcal{L}_{\mathcal{D}}\}\right\} \tag{15}$$

where $f$ and $\lambda$ are the weights of feature propagation loss and domain classification loss. $\theta_{\mathcal{F}}, \theta_y, \theta_{\mathcal{D}}$ denote the trainable parameters of GNN encoder, label propagation node classifier and conditional domain discriminator, respectively. Following (Ganin et al., 2016), a Gradient Reversal Layer (GRL) is inserted between the conditional domain discriminator and the GNN encoder to make them compete against each other during back-propagation. On one hand, $\max_{\theta_{\mathcal{D}}}\{-\mathcal{L}_{\mathcal{D}}\}$ enables the conditional domain discriminator to accurately distinguish the embeddings from different networks, conditioning on the class-label information predicted by the label propagation node classifier. On the other hand, $\min_{\theta_{\mathcal{F}}}\{-\mathcal{L}_{\mathcal{D}}\}$ encourages GNN encoder to fool conditional domain discriminator, by generating network-indistinguishable embeddings, conditioning on the predicted class-label information. After the training

**Algorithm 1: DM-GNN**

**Input**: Source network $\mathcal{G}^s = (V^s, E^s, A^s, X^s, Y^s)$, target network $\mathcal{G}^t = (V^t, E^t, A^t, X^t)$, batch size $b$, feature propagation weight $\hbar$, domain adaptation weight $\lambda$.

| | |
|---|---|
| 1 | while not max iteration do: |
| 2 |   for each mini-batch $B$ do: |
| 3 |     For each $v_i^s \in V^s$ and each $v_j^t \in V^t$ in $B$: |
| 4 |       Learn ego-embedding in (1), neighbor-embedding in (4), and final embedding in (5); |
| 5 |     end for |
| 6 |     Compute $\mathcal{L}_\mathcal{F}$ in (8) based on $\{(e_i^s, a_i^s, y_i^s)\}_{i=1}^{b/2}$ and $\{(e_j^t, a_j^t)\}_{j=1}^{b/2}$; |
| 7 |     Refine label probabilities $\{\hat{y}_i^s\}_{i=1}^{b/2}$ in (10), and compute $\mathcal{L}_\mathcal{Y}$ in (11) based on $\{(\hat{y}_i^s, y_i^s)\}_{i=1}^{b/2}$; |
| 8 |     Compute $\mathcal{L}_\mathcal{D}$ in (14) based on $\{(e_i^s, \hat{y}_i^s, d_i)\}_{i=1}^{b/2}$ and $\{(e_j^t, \hat{y}_j^t, d_j)\}_{j=1}^{b/2}$; |
| 9 |     Update parameters $\theta_\mathcal{F}, \theta_\mathcal{Y}, \theta_\mathcal{D}$ to optimize (15) via SGD; |
| 10 |   end for |
| 11 | end while |
| 12 | Use optimized $\theta_\mathcal{F}^*$ to learn $\{e_i^s\}_{i=1}^{\mathcal{N}^s}$ and $\{e_j^t\}_{j=1}^{\mathcal{N}^t}$ via (1), (4) and (5); |
| 13 | Apply optimized $\theta_\mathcal{Y}^*$ on $\{e_j^t\}_{j=1}^{\mathcal{N}^t}$ to predict $\{\hat{y}_j^t\}_{j=1}^{\mathcal{N}^t}$ via (9). |

**Output**: Cross-network embeddings: $\{e_i^s\}_{i=1}^{\mathcal{N}^s}$ and $\{e_j^t\}_{j=1}^{\mathcal{N}^t}$; Predicted node labels for $\mathcal{G}^t$: $\{\hat{y}_j^t\}_{j=1}^{\mathcal{N}^t}$.

converges, the network-invariant embeddings, which also match well with the class-conditional distributions across networks, can be learned by DM-GNN.

Algorithm 1 shows the training process of DM-GNN. It adopts a mini-batch training strategy, where half of the nodes are sampled from $\mathcal{G}^s$ and the other half from $\mathcal{G}^t$. Firstly, in each mini-batch, the cross-network embeddings are learned by dual feature extractors in Lines 3-5. The feature propagation loss is computed in Line 6. The label probabilities of the source-network nodes are refined by the label-aware propagation node classifier and the node classification loss is computed, as in Line 7. Next, by using the tensor product between the embedding vector and predicted label probability vector as the input to the conditional domain discriminator, the domain classification loss is computed in Line 8. The DM-GNN is optimized by stochastic gradient descent (SGD) in Line 9. Finally, the optimized parameters are employed to generate cross-network embeddings in Line 12. The labels of the nodes from $\mathcal{G}^t$ are predicted by the label propagation node classifier in Line 13.

*3.6 Analysis of CNNC Problem w.r.t. Domain Adaptation*

The CNNC problem can be seen as applying domain adaptation on node classification. Here, we discuss the relation between the model design of DM-GNN and the domain adaptation theory. Firstly, we follow (Ganin et al., 2016) to consider the source and target domains over the fixed representation space $e$, and a family of source classifiers $\hbar$ in hypothesis space $\mathcal{H}$. The source risk $R_s(\hbar)$ and target risk $R_t(\hbar)$ of a node classifier $\hbar \in \mathcal{H}$ w.r.t. the source distribution $\mathbb{P}_s$ and target distribution $\mathbb{Q}_t$ are given by:

$$R_s(\hbar) = \mathbb{E}_{(e,y) \sim \mathbb{P}_s}[\hbar(e) \neq y]$$

$$R_t(\hbar) = \mathbb{E}_{(e,y) \sim \mathbb{Q}_t}[\hbar(e) \neq y] \quad (16)$$

Let $\hbar^* = \arg\min_{\hbar \in \mathcal{H}} R_s(\hbar) + R_t(\hbar)$ be the joint ideal hypothesis for both $\mathbb{P}_s$ and $\mathbb{Q}_t$, the distribution discrepancy between the source and target domains is defined as (Long et al., 2018):

$$disc(\mathbb{P}_s, \mathbb{Q}_t) = |R_s(\hbar, \hbar^*) - R_t(\hbar, \hbar^*)| \tag{17}$$

where $R_s(\hbar, \hbar^*) = \mathbb{E}_{(e,y) \sim \mathbb{P}_s}[\hbar(e) \neq \hbar^*(e)]$ and $R_t(\hbar, \hbar^*) = \mathbb{E}_{(e,y) \sim \mathbb{Q}_t}[\hbar(e) \neq \hbar^*(e)]$ are the disagreement between the two hypotheses $\hbar, \hbar^* \in \mathcal{H}$ w.r.t. $\mathbb{P}_s$ and $\mathbb{Q}_t$ respectively.

According to the domain adaptation theory (Ben-David et al., 2010), the target risk $R_t(\hbar)$ is bounded by the source risk $R_s(\hbar)$ and domain discrepancy $disc(\mathbb{P}_s, \mathbb{Q}_t)$, i.e.,

$$R_t(\hbar) \leq R_s(\hbar) + disc(\mathbb{P}_s, \mathbb{Q}_t) + [R_s(\hbar^*) + R_t(\hbar^*)] \tag{18}$$

On one hand, to reduce source risk $R_s(\hbar)$, a typical solution is to minimize the classification loss on the labeled source domain (Ganin et al., 2016; Long et al., 2018), which is also adopted in DM-GNN by minimizing $\mathcal{L}_y$ in Eq. (11). In contrast to previous literature, DM-GNN proposes a label-aware propagation mechanism for the labeled source network, which includes the label-aware feature propagation loss in Eq. (7) as a part of $\mathcal{L}_\mathcal{F}$ and the label-aware label propagation mechanism in the node classifier in Eq. (10). This label-aware propagation mechanism can promote intra-class propagation while avoiding inter-class propagation based on the observed labels of the source network, thus, yielding more label-discriminative source embeddings to effectively reduce the source risk $R_s(\hbar)$.

On the other hand, to reduce domain discrepancy $disc(\mathbb{P}_s, \mathbb{Q}_t)$, DM-GNN employs the conditional adversarial domain adaptation algorithm (Long et al., 2018) by minimaxing the domain classification loss $\mathcal{L}_\mathcal{D}$ in Eq. (14). It has been theoretically proven in (Long et al., 2018) that training the optimal domain discriminator can produce an upper bound of domain discrepancy $disc(\mathbb{P}_s, \mathbb{Q}_t)$. In (Long et al., 2018), the domain discriminator takes the tensor product between the embedding vector and the own label prediction vector of each sample as the input. Instead, the proposed DM-GNN utilizes the neighborhood-refined label prediction to replace the own label prediction, which can condition adversarial domain adaptation on more accurate label prediction by taking the neighborhood information into account.

In summary, by optimizing the overall objective function of DM-GNN in Eq. (15), the target risk bounded by the source risk and domain discrepancy can be effectively reduced to achieve good node classification performance in the target network.

*3.7 Complexity Analysis of DM-GNN*

The time complexity of computing the aggregated neighbor attributes as the input of FE2 is $O\left((nnz(A^s) + nnz(A^t))\mathcal{W}\right)$, where $\mathcal{W}$ is the number of attributes, $nnz(A^s)$ and $nnz(A^t)$ denote the number of non-zero elements in $A^s$ and $A^t$, which is linear to the number of edges in two networks. The time complexity of aggregating the neighbors' label probabilities is $O\left((nnz(A^s) + nnz(A^t))\mathbb{d}\mathbb{i}\right)$, where $\mathbb{d}$ is the embedding dimensionality, and $\mathbb{i}$ is the number of training iterations, which is linear to the number of edges in the two networks. In addition, FE1, FE2, and the conditional domain discriminator are constructed as an MLP respectively, where the time complexity is linear to the number of nodes in the two networks. Thus, the overall time complexity of DM-GNN is linear to the size of the two networks.

## 4. Experiments

In this section, we empirically evaluate the performance of the proposed DM-GNN. We aim to answer the following research questions:

**RQ1**: Whether network representation learning and domain adaptation are indispensable for CNNC?

**RQ2**: How does the proposed DM-GNN perform as compared with the state-of-the-art methods?

**RQ3**: Whether the proposed model produces meaningful embedding visualizations?

**RQ4**: How are the contributions of different components in DM-GNN?

**RQ5**: How do the hyper-parameters affect the performance of DM-GNN?

### 4.1 Experimental Setup

#### 4.1.1 Datasets

Five real-world cross-network benchmark homophilic datasets constructed in (Shen et al., 2021) were used in our experiments. Blog1 and Blog2 are two online social networks, where each network captures the friendship between bloggers. Each node is associated with an attribute vector, which are the keywords extracted from the blogger's self-description. A node is associated with one label for multi-class node classification, which indicates the blogger's interested group. In addition, Citationv1, DBLPv7 and ACMv9 are three citation networks, where each network captures the citation relationship between papers. Each node is associated with an attribute vector, i.e., the sparse bag-of-words features extracted from the paper title. A node can be associated with multiple labels for multi-label node classification, which indicate the relevant research areas of the paper. Two CNNC tasks were performed between two Blog networks, and six CNNC tasks were conducted among three citation networks.

It is worth noting that the existing CNNC literature and our proposed DM-GNN mainly focus on the CNNC problem on the homophily graphs. However, it is interesting to explore the CNNC performance on the graphs with different level of homophily.

Table 2
Statistics of the networked datasets.

| Dataset | # Nodes | # Edges | Homophily Ratio | # Attributes | # Labels |
|---|---|---|---|---|---|
| Blog1 | 2300 | 33471 | 0.40 | 8189 | 6 |
| Blog2 | 2896 | 53836 | 0.40 | | |
| Citationv1 | 8935 | 15113 | 0.87 | 6775 | 5 |
| DBLPv7 | 5484 | 8130 | 0.82 | | |
| ACMv9 | 9360 | 15602 | 0.89 | | |
| Squirrel1 | 1875 | 64570 | 0.20 | 2089 | 5 |
| Squirrel2 | 3148 | 141024 | 0.20 | | |

To this end, we followed (Shen et al., 2021) to randomly extract two real-world networks from the benchmark heterophilic Squirrel dataset (Rozemberczki et al., 2021). In the Squirrel dataset, each node is an article from the English Wikipedia, each edge represents the mutual links between two articles, node attributes indicate the presence of particular nouns in the articles, and nodes are categorized into 5 classes based on the amount of their average traffic. Note that Squirrel1 and Squirrel2 are two disjoint subnetworks extracted from the original Squirrel dataset, where Squirrel1 and Squirrel2 do not share any common nodes, and there are no edges connecting nodes from Squirrel1 and Squirrel2. Two CNNC tasks can be performed between Squirrel1 and Squirrel2, by selecting one as the source network and the other as the target network.

The statistics of all datasets used in our experiments are shown in Table 2. Following (Zhu et al., 2020), we measured the homophily ratio of each graph as the fraction of intra-class edges connecting nodes with the same class label. As shown in Table 2, the three citation networks (Citationv1, DBLPv7 and ACMv9) are with high homophily, the two Blog networks (Blog1 and Blog2) are with medium homophily, while the two Wikipedia networks (Squirrel1 and Squirrel2) are with low homophily.

*4.1.2 Comparing Algorithms*

Three types of algorithms, totaled 11 algorithms, were adopted to compare with proposed DM-GNN.

Traditional Domain Adaptation: **MMD** (Gretton et al., 2007) is a simple statistical metric widely utilized to match the mean of the distributions between two domains. **DANN** (Ganin et al., 2016) inserts a GRL between the feature extractor and the domain discriminator to reduce domain discrepancy in an adversarial training manner.

Graph Neural Networks: **ANRL** (Zhang et al., 2018) leverages a SAE to reconstruct the neighbor attributes of each node and a skip-gram model to capture network structures. **SEANO** (Liang et al., 2018) designs a GNN with dual inputs hinging on the attributes of the nodes and their neighbors respectively, and dual outputs predicting node labels and node contexts respectively. **GCN** (Kipf & Welling, 2017) employs a layer-wise propagation mechanism to update each node's representation by repeatedly averaging its own representation and those of its neighbors.

Cross-network Node Classification: **NetTr** (Fang et al., 2013) projects the label propagation matrices of two networks into a common latent space. **CDNE** (Shen et al., 2021) utilizes two SAEs to respectively reconstruct the topological structures of two networks. Besides, the MMD metrics were incorporated to match the distributions between two networks. **ACDNE** (Shen, Dai, et al., 2020) employs two feature extractors to learn latent representations based on each node's own attributes and the aggregated attributes of its neighbors, and employs an adversarial domain adaptation approach to learn network-invariant representations. **UDA-GCN** (M. Wu et al., 2020) employs a dual-GCN model (Zhuang & Ma, 2018) for network representation learning and an adversarial domain adaptation approach to reduce domain discrepancy. **AdaGCN** (Dai et al., 2023) incorporates adversarial domain adaptation into GCN to learn network-invariant representations. **AdaIGCN** (Dai et al., 2023) further improves on AdaGCN by adopting an improved graph convolutional filter (Li et al., 2019).

*4.1.3 Implementation Details*

In the proposed DM-GNN[1], we set the neighborhood size as *K*=3 when measuring the topological proximities via the PPMI metric. An MLP with two hidden layers was employed to construct FE1 and FE2, where the dimensionalities were set as 512 and 128 respectively for the first and second hidden layers. The embedding dimensionality was set as $\mathbb{d} = 128$. The conditional domain discriminator was constructed by an MLP with two hidden layers, with the dimensionality of each layer as 128. We set the weight of feature propagation loss as $\beta = 0.1$ on the citation networks, $\beta = 10^{-3}$ on the Blog networks, and $\beta = 10^{-4}$ on the heterophilic Wikipedia networks. DM-GNN was trained by SGD with a momentum rate of 0.9 over shuffled mini-batches, where the batch size was set to 100. We set the initial learning rate as $\mu_0 = 0.01$ for the Blog networks, $\mu_0 = 0.02$ for the citation networks, and $\mu_0 = 0.001$ for the heterophilic Wikipedia networks. Then, we followed [19] to decay the learning rate as $\mu = \frac{\mu_0}{(1+10i)^{0.75}}$ and progressively increase the domain adaptation weight as $\lambda = \frac{2}{1+exp(-10i)} - 1$, where $i$ is the training progress linearly changing from 0 to 1.

For two traditional domain adaptation baselines which only consider node attributes, an MLP (with similar settings as FE1 and FE2 in the proposed DM-GNN) was employed for feature representation learning. To tailor the GNNs which were originally developed for a single-network scenario to CNNC, we constructed a unified network with the first $\mathcal{N}^s$ nodes from the source network and the last $\mathcal{N}^t$ nodes from the target network. Then, the unified network was employed as one input network to learn cross-network embeddings. Besides, for the CNNC baselines, we used the codes provided by the authors of the algorithms and carefully tuned their hyper-parameters to report their optimal performance. Following (Dai et al., 2023; Shen, Dai, et al., 2020; Shen et al., 2021), the Micro-F1 and Macro-F1 metrics were employed to evaluate the CNNC performance. The mean and standard deviations of the F1 scores on the homophilic graphs and the heterophilic graphs are reported in Table 3 and Table 4 respectively.

*4.2 Cross-network Node Classification Results*

*4.2.1 Performance of Traditional Domain Adaptation Methods and Graph Neural Networks (RQ1)*

On one hand, as shown in Table 3, DANN outperformed MMD in all tasks. This reflects that the adversarial domain adaptation approach yields more domain-invariant representations, as compared to the statistical approach. However, both MMD and DANN performed rather poorly in all CNNC tasks. This is because the traditional domain adaptation algorithms consider each data sample as i.i.d. during feature representation learning. The i.i.d. assumption works well in text and image data. However, it is rather unsuitable for graph-structured data, since the nodes in network are indeed not independent but related to others via complex network connections.

On the other hand, the GNNs (ANRL, SEANO and GCN), which are even originally developed for a single-network scenario,

---
[1] Our code is released at https://github.com/shenxiaocam/DM_GNN.

can significantly outperform MMD and DANN in almost all the CNNC tasks. This again demonstrates that taking network topologies into account is essential for node classification (Bhagat et al., 2011; Kipf & Welling, 2017; Perozzi et al., 2014). However, ANRL, SEANO and GCN still performed significantly worse than the CNNC algorithms (CDNE, ACDNE, AdaGCN, AdaIGCN, UDA-GCN and DM-GNN) which integrate GNN with domain adaptation. This reveals that incorporating domain adaptation to reduce domain discrepancy across networks is indeed necessary for CNNC.

Therefore, to succeed in CNNC, on one hand, employing GNNs to jointly model network structures and node attributes is indispensable. On the other hand, employing domain adaptation to mitigate distribution shift across networks is also indispensable.

*4.2.2 Performance Comparison with State-of-the-Arts (RQ2)*

As shown in Table 3, NetTr performed the worst among all the CNNC baselines. This is because NetTr only considers network structures when learning common latent features across networks. However, the target network is very structurally different from the source network. Although CDNE also utilizes the network-specific topological structures to learn embeddings for each network, in contrast to NetTr, CDNE further leverages the less network-specific node attributes to align the embeddings between two networks. By taking both network structures and node attributes into account in cross-network embedding, CDNE therefore

Table 3
Micro-F1 and macro-F1 of CNNC on the Homophilic graphs. The highest F1 value among the algorithms are shown in Boldface. (The numbers in parentheses are the standard deviations over 5 random initializations).

| $\mathcal{G}^s \to \mathcal{G}^t$ | F1 (%) | MMD | DANN | ANRL | SEANO | GCN | NetTr | CDNE | ACDNE | AdaGCN | AdaIGCN | UDA-GCN | DM-GNN |
|---|---|---|---|---|---|---|---|---|---|---|---|---|---|
| Blog1→Blog2 | Micro | 43.85 (0.4) | 44.95 (0.49) | 47.76 (1.55) | 49.87 (1.81) | 51.14 (1.37) | 50.14 (0) | 66.6 (0.64) | 66.25 (0.56) | 62.99 (1.34) | 57.16 (2.78) | 64.74 (1.13) | **66.8 (0.27)** |
| | Macro | 43.7 (0.45) | 44.84 (0.42) | 45.91 (1.52) | 49.59 (1.92) | 47.88 (2.05) | 49.18 (0) | 66.43 (0.52) | 66 (0.53) | 62.26 (1.38) | 55.66 (2.65) | 64.29 (1.26) | **66.53 (0.3)** |
| Blog2→Blog1 | Micro | 45.95 (0.63) | 46.56 (0.69) | 44.17 (1.21) | 50.23 (1.11) | 49.83 (1.8) | 52.43 (0) | 63.84 (0.82) | 63.54 (0.43) | 60.03 (2.6) | 51.11 (5.53) | 57.99 (0.61) | **65.55 (0.25)** |
| | Macro | 45.8 (0.62) | 46.42 (0.65) | 42.26 (1.68) | 49.85 (0.83) | 46.34 (2.29) | 51.51 (0) | 63.66 (0.85) | 63.51 (0.38) | 59.41 (2.53) | 49.53 (5.73) | 57.62 (0.67) | **65.42 (0.3)** |
| Citationv1→DBLPv7 | Micro | 57.01 (0.24) | 57.85 (0.21) | 66.03 (1.28) | 69.31 (1.09) | 71.24 (0.87) | 59.88 (0) | 74.15 (0.4) | 77.35 (0.77) | 76.6 (0.53) | 77.21 (0.53) | 78.25 (0.36) | **79.5 (0.13)** |
| | Macro | 53.58 (0.25) | 55.15 (0.17) | 62.78 (0.85) | 66.94 (0.99) | 68.12 (1.36) | 55.18 (0) | 71.71 (0.67) | 76.09 (0.6) | 74.83 (0.53) | 74.37 (0.87) | 77.00 (0.77) | **78.35 (0.19)** |
| DBLPv7→Citationv1 | Micro | 53.4 (0.17) | 56.27 (0.33) | 66.64 (0.83) | 71.5 (0.62) | 71.63 (0.67) | 59.11 (0) | 79.61 (0.34) | 82.09 (0.15) | 80.54 (1.24) | 82.46 (1.04) | 77.47 (1.44) | **82.91 (0.17)** |
| | Macro | 49.62 (0.28) | 54.13 (0.35) | 63.44 (0.88) | 69.54 (1.01) | 67.19 (0.73) | 55.53 (0) | 78.05 (0.29) | 80.25 (0.2) | 78.76 (1.24) | 80.67 (1.49) | 74.88 (1.47) | **81.17 (0.28)** |
| Citationv1→ACMv9 | Micro | 54.16 (0.15) | 55.53 (0.2) | 64.46 (0.78) | 67.81 (0.7) | 71.32 (0.52) | 57.75 (0) | 77.52 (0.43) | 79.56 (0.28) | 76.01 (1.73) | 77.5 (0.97) | 76.75 (0.94) | **80.46 (0.14)** |
| | Macro | 51.15 (0.15) | 53.45 (0.26) | 62.02 (0.86) | 66.25 (1.01) | 69.19 (0.6) | 53.44 (0) | 76.79 (0.35) | 78.88 (0.34) | 75.64 (1.73) | 77.46 (1.04) | 75.88 (1.02) | **80.19 (0.22)** |
| ACMv9→Citationv1 | Micro | 54.48 (0.16) | 56.73 (0.17) | 68.41 (0.67) | 72.03 (0.53) | 73.56 (1.01) | 58.81 (0) | 78.91 (0.29) | 83.27 (0.07) | 80.16 (2.43) | 83.32 (0.94) | 81.44 (0.45) | **83.92 (0.21)** |
| | Macro | 52.01 (0.26) | 54.92 (0.23) | 65.77 (0.56) | 70.29 (0.59) | 70.03 (1.51) | 55.46 (0) | 77 (0.27) | 81.66 (0.19) | 78.05 (2.43) | 81.56 (0.71) | 79.79 (0.31) | **82.25 (0.22)** |
| DBLPv7→ACMv9 | Micro | 51.43 (0.29) | 53.11 (0.36) | 63.08 (0.62) | 66.64 (0.57) | 66.83 (0.49) | 56.23 (0) | 76.59 (0.48) | 76.34 (0.67) | 73.81 (1.21) | 75.52 (2.71) | 73.10 (0.32) | **76.64 (0.65)** |
| | Macro | 46.51 (0.27) | 50.07 (0.47) | 60.19 (0.64) | 65.28 (0.74) | 62.91 (1) | 50.99 (0) | 75.91 (0.51) | 76.09 (0.46) | 71.8 (1.21) | 72.88 (7.55) | 72.17 (0.44) | **76.42 (0.78)** |
| ACMv9→DBLPv7 | Micro | 54.48 (0.29) | 55.35 (0.24) | 64.48 (0.66) | 66.13 (0.73) | 68.22 (0.86) | 56.3 (0) | 72.03 (0.21) | 76.57 (0.51) | 75.42 (1.52) | 77.2 (0.82) | 76.40 (0.66) | **78.15 (0.23)** |
| | Macro | 51.16 (0.26) | 52.49 (0.32) | 61.03 (0.69) | 63.33 (0.6) | 64.13 (1.41) | 49.8 (0) | 69.78 (0.26) | 74.31 (0.99) | 72.25 (1.52) | 73.98 (2.9) | 74.77 (0.48) | **76.43 (0.59)** |

significantly outperformed NetTr.

However, one disadvantage of CDNE is that it models network topologies and node attributes separately, i.e., employing topologies to capture the proximities between nodes within a network, while employing attributes to capture the proximities between nodes from different networks. In contrast, ACDNE unifies network topologies and node attributes in a principled way so as to jointly capture topological proximities and attributed affinity between nodes within a network and across networks. Moreover, unlike CDNE which utilizes the MMD-based statistical approach for domain adaptation, ACDNE employs a more effective adversarial learning approach. Therefore, ACDNE can achieve better overall performance than CDNE.

Next, we discuss the performance of AdaGCN, AdaIGCN and UDA-GCN. Note that AdaGCN directly employs original GCN (Kipf & Welling, 2017) for network representation learning. While both AdaIGCN and UDA-GCN employ the GCN variants to further improve the performance of GCN, thus yielding better overall performance than AdaGCN. However, AdaIGCN and UDA-GCN still performed worse than ACDNE and DM-GNN in most tasks. Since the GCN variants (Li et al., 2019), (Zhuang & Ma, 2018) adopt the typical GCN design, which mixes ego-embedding and neighbor-embedding at each convolution layer, they would easily suffer from over-smoothing and fail to capture the discrimination between connected nodes. A recent study showed that separating ego-embedding from neighbor-embedding contributes to more effective node classification, especially when connected nodes do not perfectly possess homophily (Zhu et al., 2020). Both ACDNE and DM-GNN employ dual feature extractors with different learnable parameters to separately learn ego-embedding from neighbor-embedding. FE2 focuses on capturing the attributed affinity between connected nodes within $K$ steps in a network, while FE1 captures the attributed affinity between two nodes even though they do not have any network connections. In addition, when a node has attributes which are very distinct from its neighbors, FE1 would be more effective in capturing the discrimination between the node and its neighbors. In contrast, FE2 is more advantageous when two connected nodes are perfectly satisfied with the homophily assumption. Thus, with the integration of FE1 and FE2, both commonality and discrimination can be well captured by ACDNE and DM-GNN to yield more informative representations, as compared to the CNNC baselines which adopt GCN-like models.

Moreover, one can see that DM-GNN outperformed ACDNE in all the CNNC tasks. For example, DM-GNN yielded 3% higher Micro-F1 and Macro-F1 scores than ACDNE, in the task from Blog2 to Blog1. Also, DM-GNN achieved 2% higher Micro-F1 score and 3% higher Macro-F1 score than ACDNE, in the CNNC task from ACMv9 to DBLPv7. It is worth noting that DM-GNN is distinct from ACDNE in terms of three aspects. Firstly, to enhance training stability, a feature propagation loss is incorporated into DM-GNN to make the final embedding of each node similar to a weighted average of the embeddings of its neighbors. Secondly, to refine label prediction, a label propagation mechanism is incorporated into the node classifier by DM-GNN to combine the label prediction of each node and its neighbors. In addition, a label-aware propagation mechanism is devised in DM-GNN to promote intra-class propagation while avoiding inter-class propagation for the labeled source network, which guarantees more label-discriminative source embeddings. Thirdly, to better mitigate the distribution discrepancy across networks, a conditional adversarial domain adaptation approach is employed in DM-GNN to condition adversarial domain adaptation on the neighborhood-refined class-label probabilities predicted by the label propagation node classifier. As a result, the corresponding class-conditional distributions across networks can be better matched. The outperformance of DM-GNN over ACDNE demonstrates that the incorporated feature propagation loss, label propagation node classifier and conditional domain discriminator can further boost the CNNC performance.

Table 4
Micro-F1 and macro-F1 of CNNC on the heterophilic graphs. The highest F1 value among the algorithms are shown in boldface. (The numbers in parentheses are the standard deviations over 5 random initializations).

| $\mathcal{G}^s \rightarrow \mathcal{G}^t$ | F1(%) | GCN | CDNE | ACDNE | AdaGCN | UDA-GCN | DM-GNN |
|---|---|---|---|---|---|---|---|
| Squirrel1→Squirrel2 | Micro | 20.92 (0.64) | 20.32 (0.48) | 30.84 (0.20) | 23.77 (0.28) | 23.67 (0.51) | **34.22 (0.95)** |
| | Macro | 18.25 (1.74) | 19.15 (0.95) | 30.75 (0.18) | 21.76 (0.49) | 22.06 (0.64) | **34.00 (0.81)** |
| Squirrel2→Squirrel1 | Micro | 21.07 (0.77) | 19.34 (0.79) | 31.50 (0.40) | 23.95 (0.27) | 22.30 (0.97) | **33.85 (1.00)** |
| | Macro | 18.01 (1.17) | 17.28 (0.90) | 31.42 (0.45) | 20.53 (1.38) | 21.54 (1.05) | **33.89 (0.63)** |

Finally, we report the performance of the state-of-the-art CNNC baselines and the proposed DM-GNN on the heterophilic graphs in Table 4. We can observe that CDNE performed the worst among all comparing methods on the heterophilic graphs. This is because CDNE separately models network topologies and node attributes, and learns node embeddings by reconstructing the network proximity matrix. However, the connected nodes on the heterophilic graphs tend to have different class labels, directly reconstructing the network proximity matrix of such heterophilic graphs would make the connected nodes belonging to different classes have similar node embeddings, which unavoidably degrades the node classification performance. In addition, one can see that GCN, AdaGCN, and UDA-GCN all achieved significantly lower scores than ACDNE and DM-GNN on the heterophilic graphs. We believe the reason behind is that ACDNE and DM-GNN employ dual feature extractors to learn ego-embedding and neighbor-embedding of each node separately, which is capable of capturing both commonality and discrimination between connected nodes. While GCN, AdaGCN, and UDA-GCN which employ the GCN-like models for node embedding learning would fail to capture discrimination, consequently yielding unsatisfactory CNNC performance especially on the graphs with medium homophily (i.e., Blog networks on Table 3) and heterophily (i.e., Wikipedia networks on Table 4).

*4.3 Visualization of Cross-network Embedding (RQ3)*

The t-SNE toolkit (Maaten & Hinton, 2008) was employed to visualize the cross-network embeddings learned by different algorithms. Figs. 2 and 3 show the visualization results on the Blog networks and the citation networks respectively. Firstly, DM-GNN maps the nodes from different classes into separable areas, i.e., the embeddings generated by DM-GNN are indeed label-discriminative. In addition, nodes belonging to the same class but from different networks have been mapped into the same cluster by DM-GNN, i.e., the embeddings generated by DM-GNN are network-invariant and the corresponding class-conditional distributions across networks have been well matched.

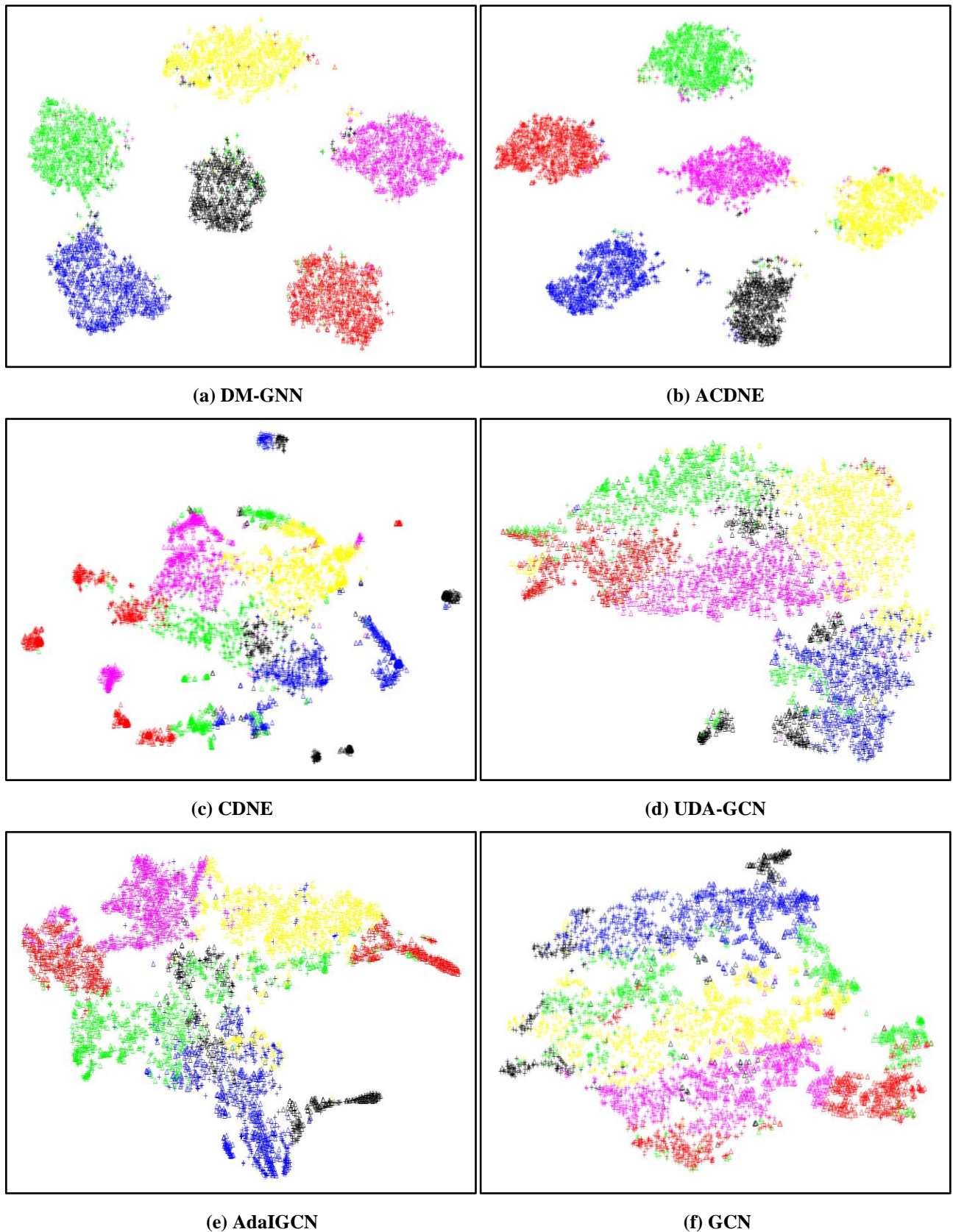

Fig. 2. Visualization of cross-network embeddings learned by different algorithms for the task from Blog1 to Blog2. Different colors are used to represent different labels. The triangle and plus symbols are utilized to represent nodes from the source and the target networks.

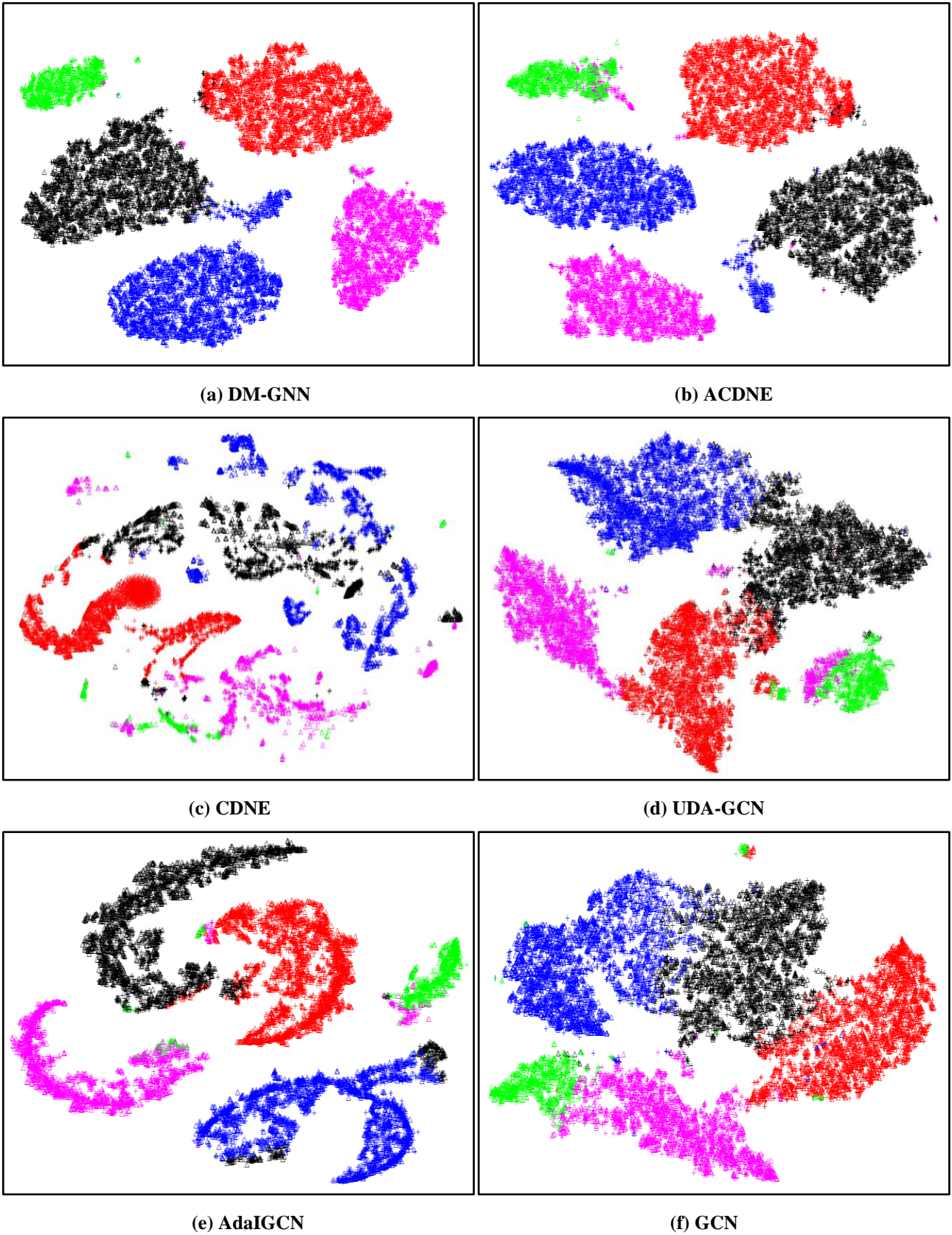

Fig. 3. Visualization of cross-network embeddings learned by different algorithms for the task from Citationv1 to DBLPv7. Different colors are used to represent different labels. The triangle and plus symbols are utilized to represent nodes from the source and the target networks.

However, the class boundaries of cross-network embeddings learned by GCN are not clear. This might be caused by the over-smoothing issue, i.e., the embeddings are over-smoothed and the nodes from different classes finally have indistinguishable embeddings after iterative layer-wise propagation (Li et al., 2018; Liu et al., 2020). In addition, although UDA-GCN and AdaIGCN utilize the GCN variants instead of the original GCN for network representation learning, they still generate much less clear class boundaries than DM-GNN. This is because the GCN variants (Li et al., 2019; Zhuang & Ma, 2018) still adopt the typical design which mixes ego-embedding and neighbor-embedding, and also entangles neighborhood aggregation and representation transformation at each layer. Unlike GCN, the GNN encoder in DM-GNN separates ego-embedding from neighbor-embedding via dual feature extractors, and also decouples neighborhood aggregation from representation transformation in FE2. Thus, DM-GNN can better preserve the individuality of each node, and avoid the nodes of different classes from becoming indistinguishable. Besides, one can see that the network representation learning ability of DM-GNN is much better than CDNE. This reflects that the joint modeling of network structures and node attributes is able to obtain more meaningful representations than modeling the two kinds of information separately.

It can be seen that both DM-GNN and ACDNE yield more label-discriminative and network-invariant embeddings than other baselines. This is because, instead of the use of GCN-like models in the baselines, both DM-GNN and ACDNE employ dual feature extractors to construct the GNN encoder so as to better preserve the discrimination between connected nodes and to alleviate the over-smoothing issue. This again confirms that an effective GNN encoder is the key success factor for CNNC. As shown in Fig. 2(a) and Fig. 2(b), for the learning task transferring from Blog1 to Blog2, DM-GNN and ACDNE have similar visualization performance, which is consistent with the results in Table 3, i.e., DM-GNN achieves comparable F1 scores with ACDNE on the Blog1-to-Blog2 task. For the learning task transferring from Citationv1 to DBLPv7, DM-GNN significantly outperforms ACDNE by an absolute 2.15% and 2.26% in terms of Micro-F1 and Macro-F1, and a consistent trend is also evident from the visualization. As shown in Fig. 3(b), for ACDNE, a small number of magenta nodes, both the triangle (i.e. source) and plus (i.e. target) symbols, are found in the cluster of green nodes. This is clearly avoided by DM-GNN, as shown in Fig. 3(a), where the magenta triangles (i.e. source) are kept apart from the green triangles (i.e. source). This reflects that DM-GNN produces more label-discriminative source embeddings, which benefits from the label-aware propagation mechanism devised for the fully labeled source network to promote intra-class propagation while avoiding inter-class propagation. On the other hand, DM-GNN further avoids the magenta plus nodes (i.e. target) from getting close to the green triangles (i.e. source). This reflects that DM-GNN can better align with the corresponding class-conditional distributions across networks, benefiting from conditional adversarial domain adaptation on the neighborhood-refined label prediction in DM-GNN.

*4.4 Ablation Study (RQ4)*

We conducted extensive ablation study to investigate the contribution from different components in DM-GNN. Five variants DM-GNN were built. The results are shown in Table 5. Firstly, the variant without FE1 yields lower F1 scores than DM-GNN since FE1 can capture the similarity of attributes between nodes even without any network connections. In addition, FE1 can also preserve discriminative ego-embeddings for connected nodes with dissimilar attributes. Thus, FE1 is a key component in the model design of DM-GNN. Secondly, the variant without FE2 leads to much lower F1 scores than DM-GNN. Since FE2 utilizes the neighbors' aggregated attributes as the input, which is essentially performing feature propagation, the necessity of feature propagation on node classification has also been verified in other GNNs (Hamilton et al., 2017; Kipf & Welling, 2017; Liang et al., 2018; Veličković et al., 2018). In fact, DM-GNN without either FE1 or FE2 has poor performance, reflecting that FE1 and FE2 can capture the complementary information to each other. Thirdly, the performance of the variant without feature propagation loss is lower than DM-GNN, which demonstrates that besides performing feature propagation in the input process, it is also helpful to ensure that the output embeddings meet the feature propagation goal. Fourthly, the variant without label propagation mechanism also has lower F1 scores than DM-GNN, indicating that incorporating the label propagation mechanism into node classifier to refine each node's label prediction can further boost the CNNC performance. Lastly, the variant without conditional domain discriminator yields significantly lower F1 scores than DM-GNN, verifying that reducing domain discrepancy across networks is necessary in CNNC.

Table 5
Micro-F1 and macro-F1 of DM-GNN variants.

| Model Variants | F1 (%) | Citationv1→ DBLPv7 | DBLPv7→ Citationv1 | Citationv1→ ACMv9 | ACMv9→ Citationv1 | DBLPv7→ ACMv9 | ACMv9→ DBLPv7 | Blog1→ Blog2 | Blog2→ Blog1 |
|---|---|---|---|---|---|---|---|---|---|
| DM-GNN | Micro | 79.50 | 82.91 | 80.46 | 83.92 | 76.64 | 78.15 | 66.80 | 65.55 |
| | Macro | 78.35 | 81.17 | 80.19 | 82.25 | 76.42 | 76.43 | 66.53 | 65.42 |
| w/o FE1 | Micro | 78.75 | 81.32 | 80.26 | 83.27 | 76.53 | 77.76 | 52.97 | 49.88 |
| | Macro | 77.79 | 79.59 | 80.31 | 81.69 | 75.90 | 75.77 | 52.27 | 48.49 |
| w/o FE2 | Micro | 70.93 | 70.97 | 68.97 | 74.44 | 63.16 | 68.70 | 42.97 | 52.29 |
| | Macro | 67.94 | 67.32 | 67.52 | 71.73 | 60.19 | 65.22 | 40.41 | 52.01 |
| w/o Feature Propagation Loss | Micro | 79.36 | 82.29 | 79.31 | 84.16 | 74.31 | 77.70 | 65.59 | 60.36 |
| | Macro | 78.16 | 80.47 | 79.00 | 82.53 | 74.00 | 76.44 | 65.34 | 59.24 |
| w/o Label Propagation | Micro | 78.85 | 82.16 | 79.15 | 83.54 | 75.55 | 76.98 | 62.87 | 64.73 |
| | Macro | 77.64 | 80.42 | 78.81 | 81.79 | 75.40 | 75.28 | 62.76 | 64.71 |
| w/o Conditional Domain Discriminator | Micro | 75.64 | 79.54 | 78.07 | 76.57 | 68.78 | 69.93 | 59.05 | 58.21 |
| | Macro | 74.03 | 77.45 | 77.65 | 74.95 | 67.63 | 67.85 | 58.54 | 57.60 |

*4.5 Parameter Sensitivity (RQ5)*

The sensitivities of hyper-parameters $f, d, K$ on the performance of DM-GNN were studied. For the weight of feature propagation loss $f$, as shown in Fig. 4(a), setting smaller value of $f$ (i.e. $10^{-3}$) for the dense Blog networks yields good performance. In contrast, setting relatively larger value of $f$ (i.e. $10^{-1}$) for the sparse citation networks would achieve good performance. For the embedding dimensionality $d$, as shown in Fig. 4(b), setting $d \in \{128, 256, 512\}$ for the Blog networks gives good performance whereas smaller values of $d$ leads to inferior performance. For the citation networks, DM-GNN is insensitive to $d$, i.e., good performance is achieved with $d \in \{32, 64, 128, 256, 512\}$.

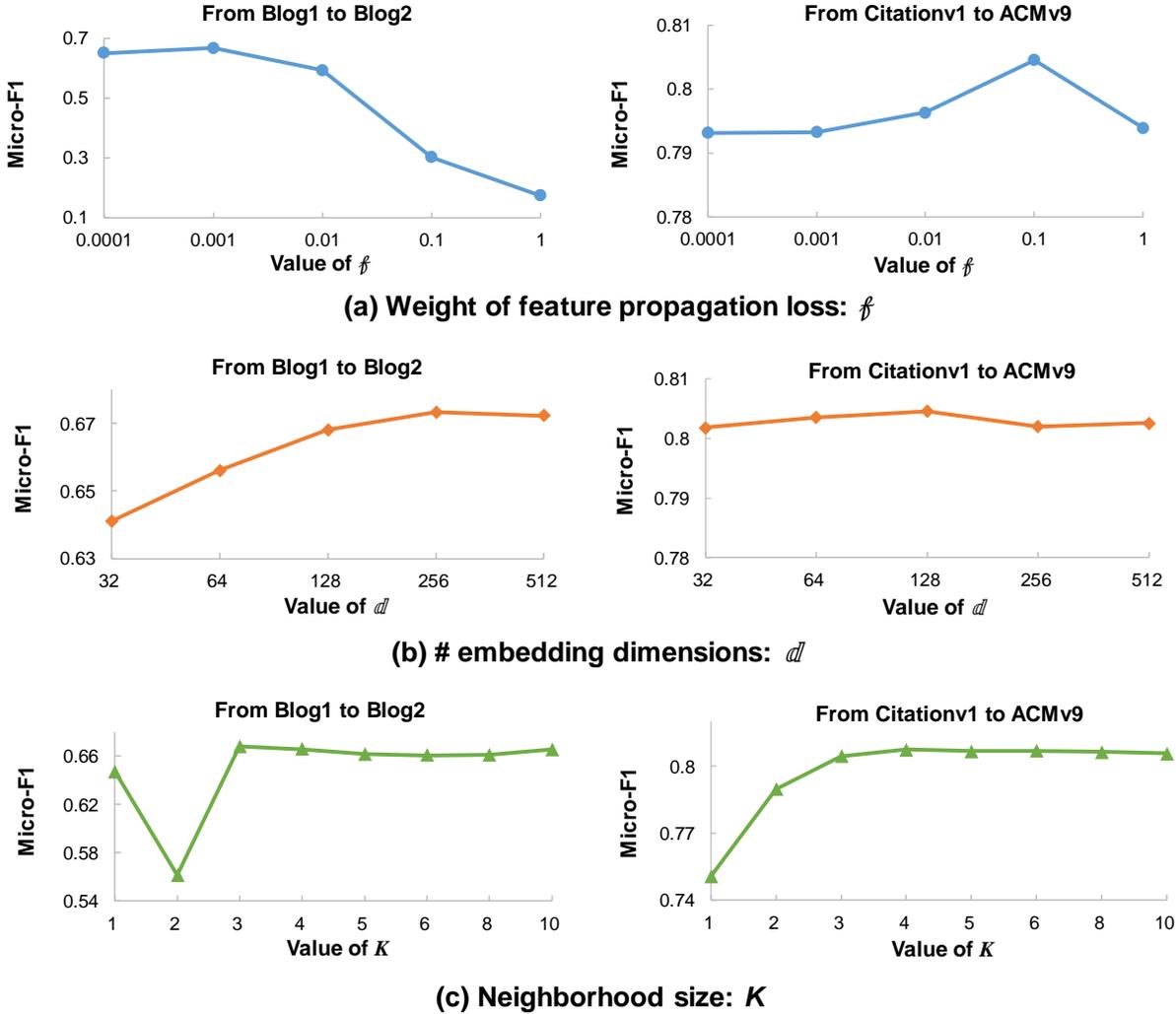

**(a) Weight of feature propagation loss:** $f$

**(b) # embedding dimensions:** $d$

**(c) Neighborhood size:** $K$

Fig. 4. Sensitivities of the hyper-parameters $f, d, K$ on the performance of DM-GNN.

For the neighborhood size $K$ on message propagation, as shown in Fig. 4(c), $K \geq 1$ in the citation networks always yields significantly higher F1 scores than $K = 1$. This reflects that taking advantage of high-order proximities is helpful for node classification, which agrees with previous works (Cao et al., 2015; Shen & Chung, 2017; Shen et al., 2021). Surprisingly, as shown in Fig. 4(c), for the learning task transferring from Blog1 to Blog2, the performance drops significantly when $K = 2$ and then rebounds to a level greater than that at $K = 1$ when $K \geq 3$. This might be due to the properties of the Blog datasets. In fact, a similar phenomenon (i.e. performance drop when $K = 2$) was also observed in CDNE (Shen et al., 2021). To investigate the phenomenon, for different $K$ values, we counted the number of connected node pairs and evaluate the proportion of connected node pairs belonging to the same class as a higher proportion of such node pairs is beneficial for node classification. As shown in Table 6, it is found that when $K=2$, the fraction of connected node pairs belonging to the same class is the lowest (i.e., 0.2850 in Blog1 and 0.2859 in Blog2) among all different $K$ values, thus leading to poor node classification performance. However, it is noted that even the proportion at $K=1$ is the highest proportion but the Micro-F1 value is significantly lower than that when $K \geq 3$. This is because when $K=1$, despite having the highest proportion, the number of connected node pairs belonging to the same class (i.e., 13359 in Blog1 and 21544 in Blog2) is much smaller than that obtained when $K \geq 3$ (i.e., more than 80000 in Blog1 and more than 130000 in Blog2). That is, the number of connected nodes pairs with common labels is rather insufficient when $K=1$. Thus, it is necessary to utilize high-order proximities by considering more distant neighbors.

Table 6
Statistics of the Blog1 and Blog2 datasets at different $K$ values

| Dataset | Statistics | $K$ | | | | | | | |
|---|---|---|---|---|---|---|---|---|---|
| | | 1 | 2 | 3 | 4 | 5 | 6 | 8 | 10 |
| Blog1 | Number of connected node pairs | 33471 | 269055 | 292936 | 313124 | 322789 | 328695 | 334581 | 337213 |
| | Number connected node pairs belonging to the same class | 13359 | 76683 | 85911 | 92214 | 95395 | 97240 | 99051 | 99811 |
| | Fraction of connected node pairs belonging to the same class | 0.3991 | **0.2850** | 0.2933 | 0.2945 | 0.2955 | 0.2958 | 0.2960 | 0.2960 |
| Blog2 | Number connected node pairs | 53836 | 428391 | 469457 | 502201 | 517304 | 526770 | 536422 | 540878 |
| | Number connected node pairs belonging to the same class | 21544 | 122481 | 138233 | 148580 | 153697 | 156752 | 159711 | 160986 |
| | Fraction of connected node pairs belonging to the same class | 0.4002 | **0.2859** | 0.2945 | 0.2959 | 0.2971 | 0.2976 | 0.2977 | 0.2976 |

## 5. Conclusions

We have proposed a novel domain-adaptive message passing GNN, named DM-GNN, which integrates GNN with conditional adversarial domain adaptation to effectively address the challenging CNNC problem. Firstly, dual feature extractors with different learnable parameters are employed to separately learn ego-embedding from neighbor-embedding of each node so as to jointly capture both the commonality and discrimination between connected nodes. Secondly, DM-GNN unifies feature propagation and label propagation in cross-network embedding, which propagates the input attributes, the output embeddings and the label prediction of each node over its neighborhood. In particular, a label-aware propagation scheme is devised for the fully labeled source network, to promote intra-class propagation while avoiding inter-class propagation. As a result, more label-discriminative source embeddings can be learned by DM-GNN. Thirdly, a conditional adversarial domain adaptation approach is employed by DM-GNN to condition GNN and domain discriminator on the neighborhood-refined class-label probabilities during adversarial domain adaptation. As a result, the class-conditional distributions across networks can be better matched to produce label-discriminative target embeddings. Extensive experiments on real-world benchmark datasets demonstrate the distinctive CNNC performance of DM-GNN over the state-of-the-art methods.

In DM-GNN, the fixed topological proximities are utilized during both feature and label propagation among the neighbors within $K$ steps. To improve the effectiveness of message passing among the neighborhood, it is promising to adopt the attention-based GNN to learn adaptive edge weights during neighborhood aggregation. Compared to previous state-of-the-art CNNC approaches using the GCN-like models for node embedding learning, the proposed DM-GNN empowered by the dual feature extractor design can better capture discrimination between connected nodes on both homophilic and heterophilic graphs. However, the feature and label propagation mechanism designed in DM-GNN is still under the homophily assumption, which cannot explicitly avoid noises from the inter-class neighbors on the target graphs with high heterophily. Thus, in order to achieve more competitive CNNC performance on the heterophilic graphs, more research is needed to design the GNN encoder specifically for the heterophilic graphs, e.g., employing positive and negative adaptive edge weights to distinguish intra-class and inter-class neighbors during both feature and label propagation.

## Acknowledgments

This work was supported in part by Hainan Provincial Natural Science Foundation of China (No. 322RC570), National Natural Science Foundation of China (No. 62102124), the Project of Strategic Importance of the Hong Kong Polytechnic University (No. 1-ZE1V), and the Research Start-up Fund of Hainan University (No. KYQD(ZR)-22016).